%% file: main.tex
\documentclass[letterpaper]{article} 
\usepackage{aaai24}  
\usepackage{times}  
\usepackage{helvet}  
\usepackage{courier}  
\usepackage[hyphens]{url}  
\usepackage{graphicx} 
\usepackage{natbib}  
\usepackage{caption} 
\usepackage{algorithm}
\usepackage{algpseudocode}
\usepackage[accsupp]{axessibility}
\usepackage{makecell}
\usepackage{enumitem}
\usepackage{subcaption}
\usepackage{bm}
\usepackage{amssymb}

\usepackage{comment}
\usepackage{amsmath}
\usepackage{color,xcolor,colortbl}
\usepackage{epsfig}
\usepackage{booktabs}
\usepackage{multirow}
\usepackage{array}
\usepackage{adjustbox}
\usepackage{arydshln}
\usepackage{autobreak}
\usepackage{ragged2e}
\usepackage{scalerel}
\usepackage{amsthm}
\usepackage{mathtools}
\usepackage{lipsum}
\usepackage{cuted}
\usepackage{siunitx}

\newcommand{\black}[1]{\textbf{\textcolor{black}{#1}}}

\definecolor{mygray}{gray}{.9}

\newcommand{\vect}[1]{\boldsymbol{\mathbf{#1}}}

\usepackage{newfloat}
\usepackage{listings}
\DeclareCaptionStyle{ruled}{labelfont=normalfont,labelsep=colon,strut=off} 
\floatstyle{ruled}
\newfloat{listing}{tb}{lst}{}
\floatname{listing}{Listing}
\title{Adaptive Discovering and Merging for Incremental Novel Class Discovery}
\author{
    Guangyao Chen\textsuperscript{\rm 1},
    Peixi Peng\textsuperscript{\rm 2, \rm 3$\ast$},
    Yangru Huang\textsuperscript{\rm 1},
    Mengyue Geng\textsuperscript{\rm 1},
    Yonghong Tian\textsuperscript{\rm 1, \rm 2, \rm 3}\thanks{Corresponding Author.}
}
\affiliations{
    \textsuperscript{\rm 1} National Key Laboratory for Multimedia Information Processing, School of Computer Science, Peking University, China. \\
    \textsuperscript{\rm 2} School of Electronic and Computer Engineering, Shenzhen Graduate School, Peking University, China. \\
    \textsuperscript{\rm 3} Peng Cheng Laboratory, China. \\

    \{gy.chen, pxpeng, yrhuang, mygeng, yhtian\}@pku.edu.cn
}

\usepackage{bibentry}

\begin{document}

\maketitle

\begin{abstract}
One important desideratum of lifelong learning aims to discover novel classes from unlabelled data in a continuous manner. The central challenge is twofold: discovering and learning novel classes while mitigating the issue of catastrophic forgetting of established knowledge. To this end, we introduce a new paradigm called Adaptive Discovering and Merging (ADM) to discover novel categories adaptively in the incremental stage and integrate novel knowledge into the model without affecting the original knowledge. To discover novel classes adaptively, we decouple representation learning and novel class discovery, and use Triple Comparison (TC) and Probability Regularization (PR) to constrain the probability discrepancy and diversity for adaptive category assignment. To merge the learned novel knowledge adaptively, we propose a hybrid structure with base and novel branches named Adaptive Model Merging (AMM), which reduces the interference of the novel branch on the old classes to preserve the previous knowledge, and merges the novel branch to the base model without performance loss and parameter growth. Extensive experiments on several datasets show that ADM significantly outperforms existing class-incremental Novel Class Discovery (class-iNCD) approaches. Moreover, our AMM also benefits the class-incremental Learning (class-IL) task by alleviating the catastrophic forgetting problem.
\end{abstract}

\input{Main/Tex/01_introduction}

\input{Main/Tex/02_related_works}

\input{Main/Tex/03_methodology}

\input{Main/Tex/04_experiments}
\input{Main/Tex/05_conclusion}

\section{Acknowledgements}
The study was funded by the National Natural Science Foundation of China under contracts No. 62332002, No. 62027804, No. 62088102, No. 61825101, No. 62372010,  and No. 62202010, and the major key project of the Peng Cheng Laboratory (PCL2021A13). Computing support was provided by Pengcheng Cloudbrain.

\bibliography{main}

\end{document}

%% file: Main/Tex/01_introduction.tex
\section{Introduction}
\label{sec:intro}

A long-standing goal of machine learning is to build Artificial Intelligence (AI) systems that can mimic human-level performance in an open and online manner.
Neural networks, originally inspired by human brain structures, should also be flexible enough to incrementally absorb novel concepts (or \textit{classes}) after acquiring previous knowledge.
To evaluate the ability of novel class discovery and memory in the open world, several tasks~\cite{han2020automatically,han2019learning,zhong2021neighborhood,zhong2021openmix,fini2021unified,roy2022class} are proposed to continuously identify new classes while leveraging some previously learned knowledge. Among these task settings, the class-incremental Novel Class Discovery (class-iNCD)~\cite{roy2022class} aims at discovering novel concepts sequentially in a continuous manner. Compared with other settings, models in class-iNCD can not access data in previously seen categories, while still having to perform well across all classes.

This paper focuses on developing efficient neural network models for the class-iNCD task. To achieve continuous learning without revisiting past knowledge, two significant challenges emerge as paramount. The first pertains to the discovery of new categories from unlabeled data, requiring the model to discern and learn novel patterns without explicit annotations. The second challenge is the notorious issue of catastrophic forgetting. As the model learns new concepts, it risks overwriting or diminishing prior knowledge.

\begin{figure}[!t]
    \centering
    \includegraphics[width=\linewidth]{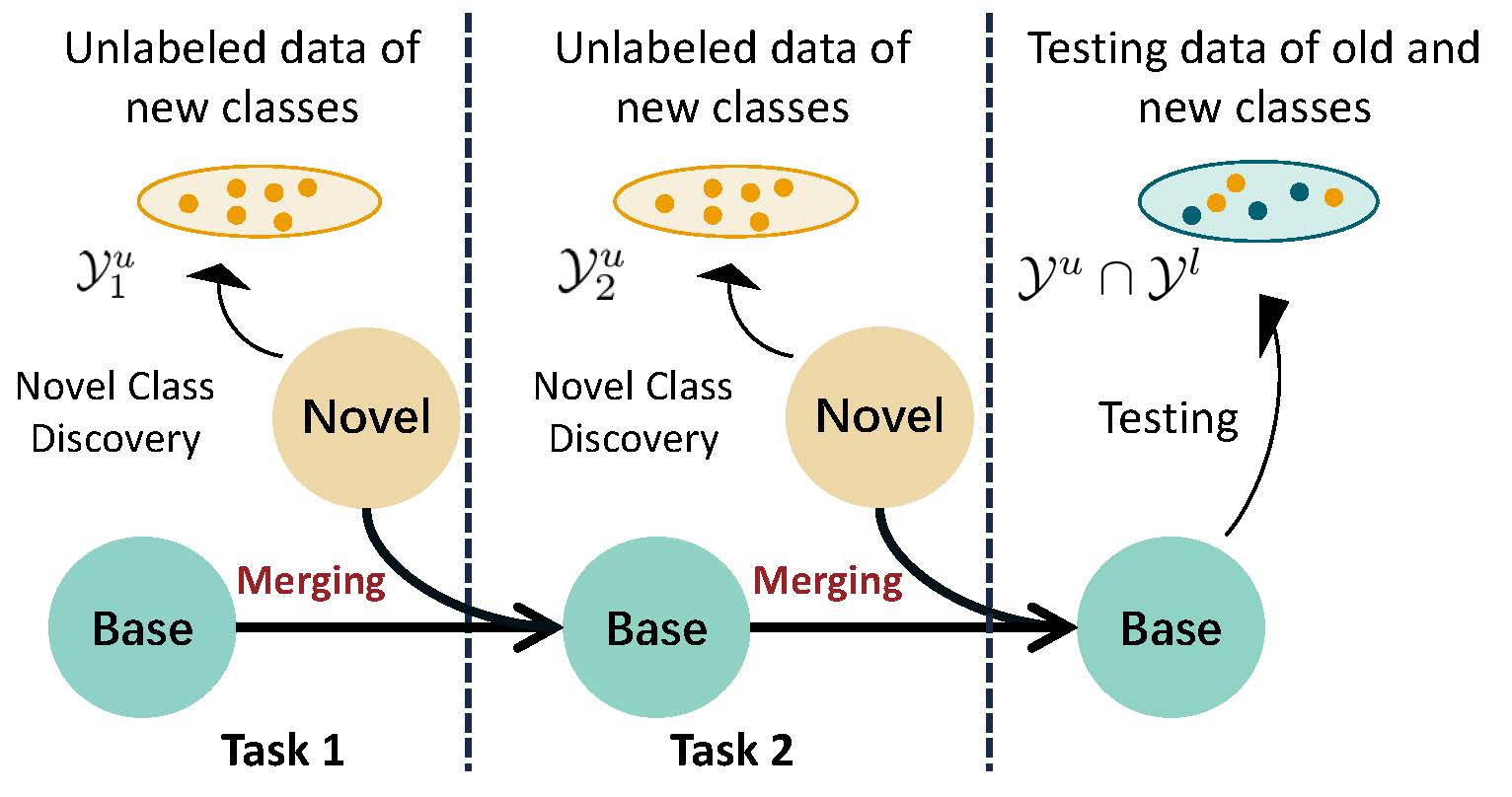}
    \caption{Illustration of incremental novel class discovery with an elastic learner. The novel model adaptively discovers novel classes from the unlabeled data of the current task based on the previous base model. The novel learner should not only affect the knowledge of the base model but also integrate with the base model.}
    \label{fig:increment}
\end{figure}

To discover new categories and learn from unlabeled data, most existing methods~\cite{han2020automatically,zhong2021neighborhood,zhong2021openmix,jia2021joint,fini2021unified} rely on pseudo-labels for both data annotation and model supervision.
However, the quality of the pseudo-label greatly affects the performance of novel category discovery. 
To address this issue, we propose to separate representation learning and category discovery in the incremental novel category discovery stage. 
Specifically, the model learns representations directly from unlabeled data through contrastive learning, while the classifiers are optimized independently with adaptive category discovery. 
To further enhance the quality of pseudo-label generation and cultivate a robust classifier, we introduce a Triplet Comparison (TC) technique to minimize intra-class distance and maximize inter-class distance. Additionally, a Probability Regularization (PR) technique is also developed to prevent the model from becoming trapped in a local optimal solution specific to a single class. By constraining the contrast and diversity of probability space, the learner could achieve self-adaptive novel category discovery from unlabeled data.

The class-iNCD also faces the well-known catastrophic forgetting problem~\cite{mccloskey1989catastrophic}. 
Several architectural approaches~\cite{rusu2016progressive, liu2021adaptive, yan2021dynamically} address this problem by adding multiple new network components for novel data, but increase the model parameters and computations with each sequential task.
Alternatively, a basic method of Incremental Model Merging (IMM) without parameter growth can be built based on model re-parameterization~\cite{zagoruyko2017diracnets,ding2019acnet,cao2020conv,guo2020expandnets,ding2021repvgg}. 
Specifically, a typical IMM method consists of a base branch and a novel branch, where the novel branch could be merged into the base model according to the additivity of convolution. 
However, IMM may disturb the base model as the novel branch is learned directly from the novel data. 
To reduce interference with the base model, we propose an Adaptive Feature Fusion (AFF) mechanism, which leverages the incompatibility of feature magnitude between old and new input of a network branch~\cite{dhamija2018reducing,chen2020learning,chen2021adversarial}.
Specifically, AFF uses a gated dynamic unit to maintain the important information of the feature from the base branch. 
In this way, the novel branch trained by AFF has a low magnitude on the old classes, and brings little influence to the base branch. 
Since AFF is performed on the fusion of features of the base branch and novel branch rather than their model parameters, we further improve the AFF to Adaptive Model Merging (AMM), where we alternatively approximate the dynamic gating information through the weight information of $\mathrm{BN}$ in the base branch to merge the novel branch to the base model.

Capitalizing on both adaptive novel discovery and AMM, our solution forms a new paradigm called Adaptive Discovering and Merging (ADM). As shown in Fig.~\ref{fig:increment}, ADM could discover novel categories adaptively in the incremental stage and integrate novel knowledge into the model without affecting the original knowledge, thereby yielding a marked enhancement in the model's efficacy.
To summarize, the main contributions of our paper are as follows: \textbf{(1)} We decouple representation learning and adaptive discovery, and propose TC and PR mechanisms for adaptive novel category assignment by constraining the probability difference and diversity. \textbf{(2)} We propose two novel AFF and AMM methods to effectively handle the catastrophic forgetting problem, especially the latter, which can make the novel branch mergeable without additional model parameters through sequential learning tasks. \textbf{(3)} Extensive experimental results show that our method not only achieves state-of-the-art performance on several class-iNCD datasets but also has a positive impact on the class-IL task.

%% file: Main/Tex/02_related_works.tex
\section{Related Works}
\label{sec:relat}

\subsubsection{Novel Class Discovery}
Novel Class Discovery (NCD) deals with the task of learning to discover novel classes in an unlabelled dataset by utilizing the knowledge acquired from another labeled base dataset~\cite{han2019learning}. 
It is assumed that the classes in the labeled and unlabelled sets are disjoint. 
Several NCD methods use a stage-wise training scheme where the model is pre-trained on the labeled base dataset, followed by fine-tuning on the unlabelled data using an unsupervised clustering loss~\cite{hsu2017learning,hsu2019multi,han2019learning,liu2022residual}. Barring~\cite{liu2022residual}, none of the above methods consider tackling the forgetting issue, and as a result, the model loses the ability to classify the base classes. 
The second category comprises NCD methods that assume both the labeled and unlabelled data are available simultaneously, which are then trained jointly~\cite{han2020automatically,zhong2021neighborhood,zhong2021openmix,jia2021joint,fini2021unified}.
However, access to seen classes limits the practical applicability of NCD methods.
Class-incremental novel class discovers (class-iNCD) \cite{roy2022class} is proposed to preserve the ability of the model to recognize previously seen base categories.
Compared with class-incremental learning, the novel data of class-iNCD at every stage are not provided with any supervised labels.

\subsubsection{Incremental Learning with Multi-branch Model}
Incremental Learning (IL) is a learning paradigm where a model is trained on a sequence of tasks such that data from only the current task is available for training, while the model is evaluated on all the observed tasks. 
To prevent catastrophic forgetting~\cite{mccloskey1989catastrophic,goodfellow2013empirical}, existing class-IL approaches with multi-branch models could be categorized into knowledge distillation and architectural methods. 
Knowledge distillation methods~\cite{jung2016less,li2017learning,kang2022class,wang2022foster} achieve better generalization performance by transferring knowledge from a base model to a novel network through matching logits, activations, attention, and so on.
Most architectural methods~\cite{rusu2016progressive, liu2021adaptive, yan2021dynamically} adjust network capacity dynamically to handle a sequence of incoming tasks.
Some methods even dedicate different model parameters for different incremental phases, to prevent model forgetting (caused by parameter overwritten).
However, these approaches require creating and storing additional network components and performing multiple forward-pass computations for inference, which incur extra computational costs.
Different from these architectural methods, our approach does not continuously increase the network size and even could achieve non-destructive incremental fusion, so as to be used as a plug-and-play unit without extra inference computational costs.

\begin{figure*}[!htbp]
    \centering
    \includegraphics[width=\linewidth]{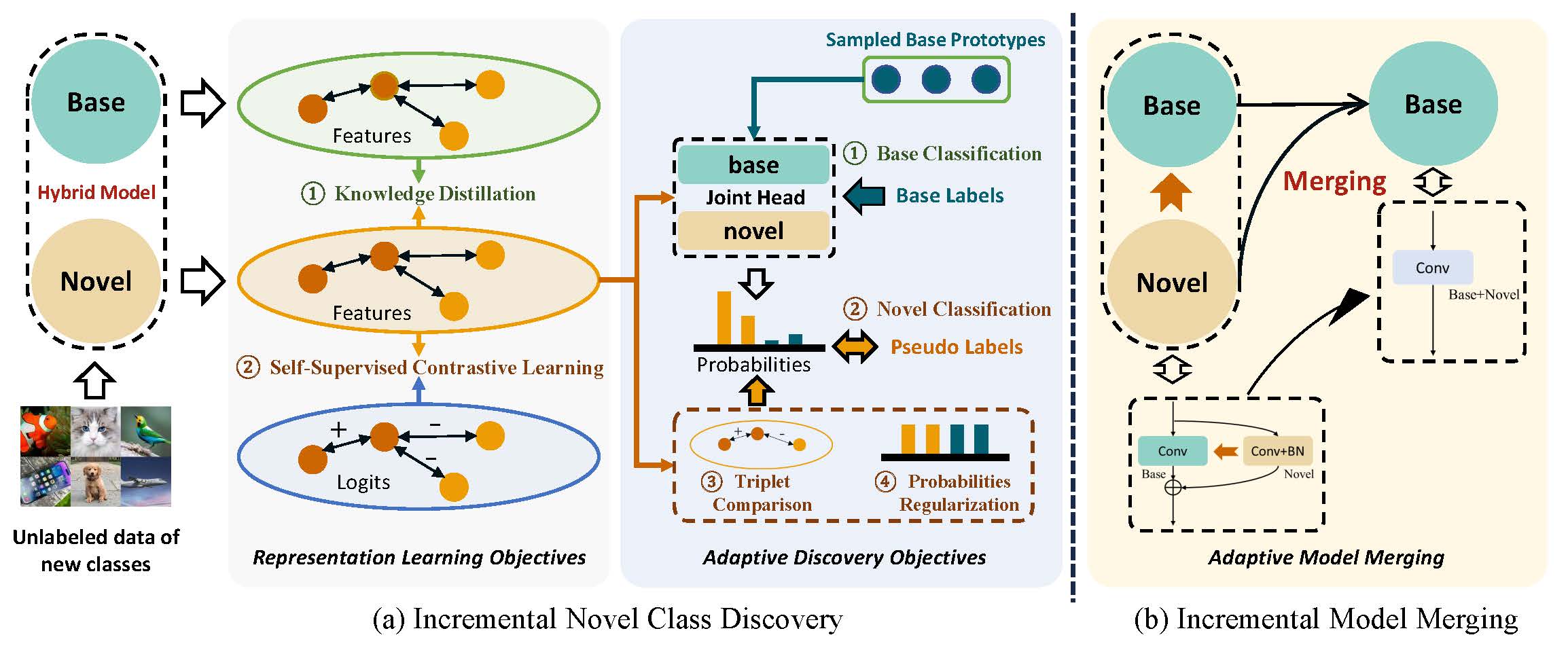}
    \caption{
    Illustration of incremental novel class discovery with adaptive discovering and merging. (a) For incremental novel class discovery, representation learning and feature-based adaptive novel class discovery are decoupled. Knowledge distillation is applied for representation learning to mitigate catastrophic forgetting, and self-supervised contrastive learning is employed to learn new categories. For adaptive novel discovery, the base head is trained with replayed base prototypes. For the category assignment of new classes, triplet comparison enhances the probability differentiation and probability regularization promotes the diversity to avoid overfitting of categories within a single class. Finally, pseudo labels are generated by the maximum probability.
    (b) Sketch of adaptive model merging. Each $\mathrm{Conv}+\mathrm{BN}$ unit is decomposed into the base and novel branch, which could be merged based on the additivity of convolution.
    }
    \label{fig:framework}
\end{figure*}

%% file: Main/Tex/03_methodology.tex
\section{Proposed Approach}
\label{sec:method}

\subsubsection{Problem Setting}
Here, we follow the setting of class-incremental Novel Class Discovery (class-iNCD) ~\cite{roy2022class}, where the lifelong learner not only distinguishes the novel classes but also preserves the performance on the previously seen classes, without having access to or storing data from the previous tasks.
In the class-iNCD, a initial supervised task contains $n^l$ instances of a labelled dataset $\mathcal{D}_{0} = \{(x^l_i, y^l_i)\}^{n}_{i=1}$, where $x^l \in \mathcal{X}^l$ represents the input data and $y^l \in \mathcal{Y}^l$ as corresponding labels.
The following task $t$ only have the unlabelled dataset $\mathcal{D}^u_t = \{(x^u_i)\}^{n}_{i=1}$, where $x^u \in \mathcal{X}^u$ are the unlabelled new data containing $c^u$ classes. Here, it is assumed that the labels in $\mathcal{Y}^l$ and $\mathcal{Y}^u$ are disjoint, \emph{i.e}, $\mathcal{Y}^l \cap \mathcal{Y}^u = \emptyset$.
The class-iNCD focuses on clustering the data in $\mathcal{D}^u_t$ by just leveraging the learned information contained in the single model, while still behaving well on the previous tasks. 
For the evaluation of the class-iNCD, a joint (task-agnostic) head is adopted to distinguish the old and new classes by directly comparing the predictions with these re-assigned ground-truth labels. Here, HA~\cite{kuhn1955hungarian} is used to re-assign ground-truth IDs based on the predictions and ground-truth labels for the new classes only. 

\subsubsection{Overview of ADM}
The key challenge to solving class-iNCD is to learn unseen classes from unlabeled data and keep the distinguish for the seen classes.
To circumvent this problem, we propose ADM, a novel paradigm that adaptively discovers and integrates new categories in the incremental stage without forgetting the original knowledge. As shown in Fig.~\ref{fig:framework}, ADM consists of two stages: incremental novel class discovery and incremental model merging. The former adaptively discovers novel classes by constraining the probability differences and diversity, and the latter merges the novel branch to the base model without performance loss and parameter growth. In the following sections, we describe the adaptive process of incremental novel class discovery and the network architectures for non-destructive incremental merging.

\subsection{Incremental Novel Class Discovery}
\label{sec:INCD}

\subsubsection{Representation Learning Objectives}
Unlike most novel class discovery settings~\cite{hsu2017learning,hsu2019multi,han2019learning,liu2022residual}, class-iNCD does not have any labeled data and only provides unlabeled data of new classes.
Therefore, our representation learning adopts self-supervised contrastive learning~\cite{chen2020simple} on all samples. 

\noindent $\bullet \ $\textit{\textbf{Novel Representation Learning.}}
Formally, given two views (random augmentations) $\{x^u_i,\hat{x}^u_i\}$ of the same image in a mini-batch $B$, the self-supervised contrastive loss is defined as:
\begin{equation}
\mathcal{L}^{r}_c=-\frac{1}{|B|} \sum_{i \in B}\log \frac{\exp \left(\boldsymbol{z}_i^\top \hat{\boldsymbol{z}}_i / \tau_u\right)}{\sum_{j \neq i} \exp \left(\boldsymbol{z}_i^\top \hat{\boldsymbol{z}}_j / \tau_u\right)}\,
\end{equation}
where the feature $\boldsymbol{z}_i=g\left(f_n\left(\boldsymbol{x}_i\right)\right)$ and is $\ell_2$-normalised, $f_n, g$ denote the backbone of novel model and the projection head, and $\tau_u$ is a temperature value.

\noindent $\bullet \ $\textit{\textbf{Base Representation Learning.}}
To further mitigate forgetting on the base classes from representation learning, we follow the FRoST~\cite{roy2022class} by using knowledge distillation~\cite{hinton2015distilling,li2017learning,roy2022class} on the novel model to ensure the feature encoding for the old task does not drift too far while learning on the new task, which is formalized as:
\begin{equation}
\label{eqn:feat-KD}
    \mathcal{L}^{r}_{\mathrm{KD}} = - \mathbb{E}_{p(x^u)} \Big|\Big|f_b(x^u) - f_n(x^u)\Big|\Big|_2,
\end{equation}
where $f_b$ is the base feature extractor from the previous task and is kept frozen. The overall objective for representation learning is given as:
$\mathcal{L}^{r}=\mathcal{L}^{r}_{\mathrm{KD}} + \mathcal{L}^{r}_c$.

\subsubsection{Adaptive Discovery Objectives}
Besides learning generic feature representations to facilitate the discovery of novel categories, another key challenge is to generate pseudo clusters/labels for unlabelled data to guide the learning of a classifier. Moreover, at any time during the training sessions, a single classification head is maintained for all the classes seen for the class-iNCD. Thus, in the novel class discovery phase, the classifier extends a novel head based on the base head to form a joint classifier.

\noindent $\bullet \ $\textit{\textbf{Novel Classification.}}
To learn novel classification, pseudo-labels are computed from novel probabilities $p^u$ to train the joint classifier $\mathcal{H}$. 
Specifically, we employ the \textit{Logit} output $\mathcal{P}^u$ of novel head $\mathcal{H}_n$ to compute the pseudo-label $\hat{y}^u$ for an unlabelled data $x^u$.  The $\hat{y}^u$ is then used to supervise the training of $\mathcal{H}$. The self-training for novel classification is formalized as:
\begin{equation}\small
\label{eqn:self-train}
    \mathcal{L}_\mathrm{self} = - \mathbb{E}_{(x^u, \hat{y}^u)} \frac{1}{|C^A|} \sum^{|C^A|}_{k=1} \hat{y}^u_k \log \sigma_k (\mathcal{H}(f_n(x^u))),
\end{equation}
where $C^A$ in Eq.3 represents the total number of categories that are configured and $\hat{y}^u = C_\mathrm{base} + \underset{k \in C_\mathrm{novel}}{{\arg\max} \, \mathcal{H}_n(f_n(x^u))}$. This part of the objective generates pseudo-labels for the unlabeled data to guide the training of the joint classifier. 

\noindent $\bullet \ $\textit{\textbf{Triplet Comparison.}}
However, pseudo-labels generated by unconstrained probabilities are inaccurate.
Therefore, a similarity prediction function parameterized by a linear novel classifier should be learned such that instances from different classes are grouped into multisets. 
To achieve this, we reformulate the cluster learning problem as a triplet similarity prediction task. 
Specifically, to obtain the triplet instances for the unlabeled set, we compute the cosine distance between all pairs of feature representations $f(x_i)$ in a mini-batch. 
We then rank the computed distances and for each instance generate the triplet instances for its most similar and dissimilar neighbors, to generate pseudo-labels from the most confident positive pair $\{\mathcal{P}^u_a, \mathcal{P}^u_p\}$ and negative pairs $\{\mathcal{P}^u_a, \mathcal{P}^u_n\}$ for each instance within the mini-batch.
The objective for triplet comparison is designed as a refined version of the traditional mean absolute error loss, combining two modifications for greater effectiveness:
\begin{equation}\small \label{eq:triplet}
    \mathcal{L}_{triplet} = \frac{1}{C^A}\sum(\sigma(\mathcal{P}^u_a ) - \sigma(\mathcal{P}^u_p ))^2 - \frac{1}{C^A}\sum(\sigma(\mathcal{P}^u_a ) - \sigma(\mathcal{P}^u_n ))^2.
\end{equation}
Here, $\sigma$ denotes the softmax function which assigns instances to one of the novel classes, and $m$ is a nonnegative margin representing the minimum difference between the positive and negative distances and set as $1.0$.  
By controlling the intra-class and inter-class variance of novel classes, the triplet comparison improves the quality of the pseudo-labels.

\noindent $\bullet \ $\textit{\textbf{Probabilities Regularization.}}
In the early stages of the training, the network could degenerate to a trivial solution in which all instances are assigned to a single class.
To prevent this solution, we introduce a regularization term for the probabilities by applying a maximum entropy regularization that regularizes the mean of all probabilities of novel classes:
\begin{equation}\small \label{eq:reg}
\mathcal{L}_{r} = -\frac{1}{|B|} \sum_{x_i \in \mathcal{D}^u_t }\sigma(\mathcal{H}_n(f_n(x_i))) \log (\frac{1}{|B|} \sum_{x_i \in \mathcal{D}^u_t} \sigma(\mathcal{H}_n(f_n(x_i))).
\end{equation}
Here, maximum entropy regularization for the mean of all probabilities of novel classes encourages diversity to avoid the overfitting of categories within a single class.

\noindent $\bullet \ $\textit{\textbf{Base Classification.}}
To further mitigate forgetting on the base classes for the classifier, we follow the FRoST~\cite{roy2022class} by using \textit{generative feature-replay} drawn from a Gaussian distribution of base feature prototypes $\bm{p}_c$ is used to preserve the performance of old classes.
While learning on the new task $t$; the weights of the joint classifier $\mathcal{H}$, corresponding to the base classes $C_\mathrm{base}$, are trained by replaying features from the class-specific Gaussian distribution $\mathcal{N}(\bm{\mu}_c, {\bm{v}_c}^2)$. The feature-replay loss is formalized as:
\begin{equation}
\label{eqn:replay}
    \mathcal{L}^c_\mathrm{base} = - \mathbb{E}_{c \sim C} \mathbb{E}_{(z, y_c) \sim \mathcal{N}(\bm{\mu}_c, {\bm{v}_c}^2)} \sum^{|C_\mathrm{base}|}_{k=1} y_{kc} \log \sigma_k(\mathcal{H}^A(p)).
\end{equation}
Then, the loss for discovering novel classes and having a single classifier for all the classes seen so far can be formalized as $\mathcal{L}^c = \mathcal{L}^c_\mathrm{base} + \{ \mathcal{L}^c_\mathrm{triplet} +   \mathcal{L}^c_\mathrm{r} +  \omega(t) \mathcal{L}^c_\mathrm{self} \}$,
where $\omega(t)$ are ramp-up functions to ensure stability in learning.
Finally, the total objective function is $\mathcal{L} = \mathcal{L}^r + \mathcal{L}^c$, which consists of two components: the representation learning loss $\mathcal{L}^r$ and the classifier loss $\mathcal{L}^c$. 

\begin{figure*}[!tp]
    \centering
    \includegraphics[width=0.85\linewidth]{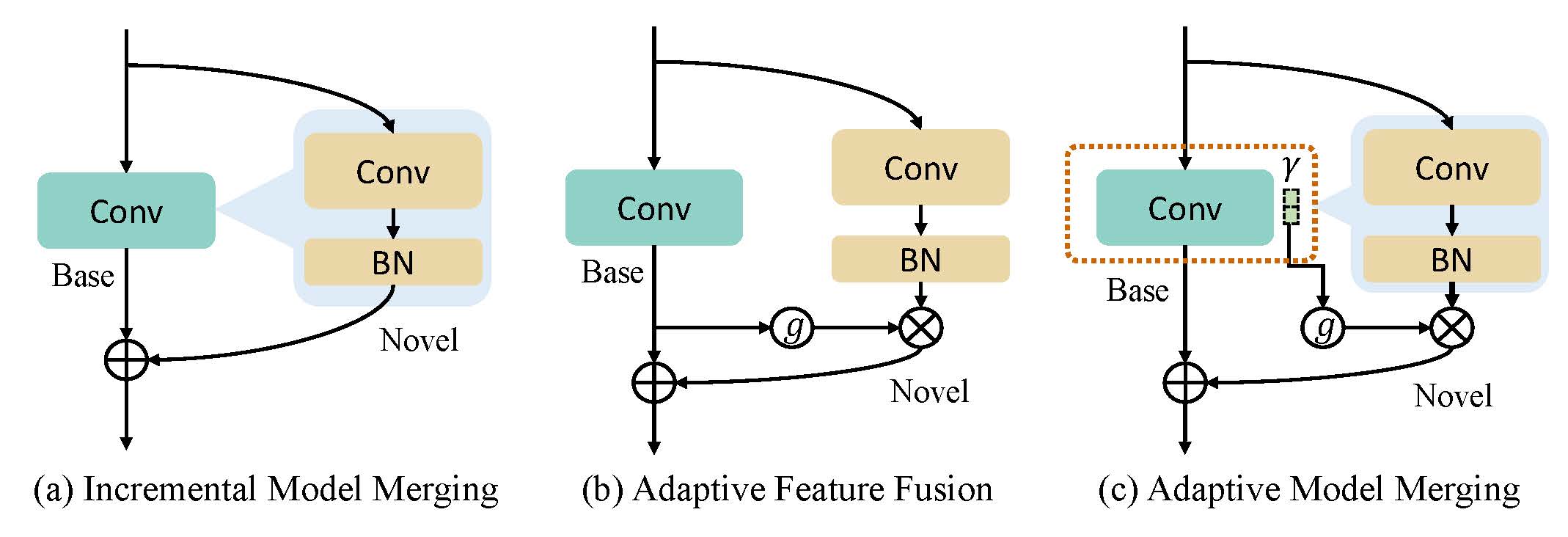}
    \caption{Conceptual illustrations of different merging methods. (a) Incremental Model Merging (IMM)~\cite{zhu2022self} decomposes each $\mathrm{CONV}+\mathrm{BN}$ operation into the base and novel parts, then fuses these two parts based on the additivity of convolution. (b) Adaptive Feature Fusion (AFF) adopts the gated output of the base branch to guide the learning process and outputs of the novel branch. (c) Adaptive Model Merging (AMM) uses the weight of base $\mathrm{BN}$ to replace dynamic gated output, so as to achieve linear merging.}
    \label{fig:structure}
\end{figure*}

\subsection{Incremental Model Merging}
\label{sec:IMM}

\noindent
To further integrate novel knowledge with the base knowledge, we further proposed a two-branch hybrid architecture that could make an only-novel-training ensemble of numerous models.
In detail, the base approach of incremental model merging is first described, which could achieve a non-destructive incremental merging mechanism.
Second, adaptive feature fusion is proposed to guide the novel branch to generate a lower magnitude to old categories and fuse two branches at the feature level.
Then we further proposed adaptive model merging to approximate the dynamic gating output through the weight of BN in the base branch, and achieve linear additional fusion in the model level.

\subsubsection{Incremental Model Merging}
Since the multi-branch topology has drawbacks for inference but the base branch seems beneficial to novel discovery and learning, we employ base and novel branches  to make an \textit{only-novel-training} ensemble of numerous models as shown in Fig.~\ref{fig:structure}(a).
Based on the homogeneity of convolution~\cite{ding2019acnet,ding2021repvgg}, each $\mathrm{CONV}+\mathrm{BN}$ unit the base model trained on labeled dataset $\mathcal{D}^l$ is fused to a single convolution layer with weight $\mathrm{W}^\prime_{(b)}$ and bias $\mathrm{B}^{\prime}_{(b)} $.
Then, for the stage of novel class discovery, we extend a novel branch with a $\mathrm{CONV}+\mathrm{BN}$ unit, where $\mathrm{W}_{(n)}^k\in\mathbb{R}^{C_2\times C_1}$ for the kernel with the same size of $\mathrm{W}^\prime_{(b)}$, and $\vect{\mu}_{(n)},\vect{\sigma}_{(n)},\vect{\gamma}_{(n)},\vect{\beta}_{(n)}$ for the $\mathrm{BN}$ following the novel convolution.
For input $\mathrm{X}^{(\text{1})}$ and output $\mathrm{X}^{(\text{2})}$, we have
\begin{equation}
    \begin{aligned}
        \mathrm{X}^{(\text{2})} &= (\mathrm{X}^{(\text{1})} \ast \mathrm{W}^\prime_{(b)}) + \mathrm{B}^\prime_{(b)} \\
        & + \mathrm{BN}(\mathrm{X}^{(\text{1})} \ast \mathrm{W}_{(n)},\vect{\mu}_{(n)},\vect{\sigma}_{(n)},\vect{\gamma}_{(n)},\vect{\beta}_{(n)}).
    \end{aligned}
	\label{Eqn:convnovel}
\end{equation}
In the novel categories training stage, the base branch parameters $\{\mathrm{W}^\prime_{(b)},\mathrm{B}^{\prime}_{(b)} \}$ are fixed, and only the novel branch parameters $ \{\mathrm{W}_{(n)},\vect{\mu}_{(n)},\vect{\sigma}_{(n)},\vect{\gamma}_{(n)},\vect{\beta}_{(n)} \} $ is trained, where the novel output is added to the base output, so as to integrate the base knowledge.

After nove class discovery, we will have two $k \times k$ kernels, and two bias vectors. 
Then we obtain the final base kernel $\mathrm{W}^\prime = \mathrm{W}^\prime_{(b)} + \mathrm{W}^\prime_{(n)}$ and bias $\mathrm{B}^\prime = \mathrm{B}^\prime_{(b)} + \mathrm{B}^\prime_{(n)}$ by adding up the two kernels and two bias vectors.
For input $\mathrm{X}^{(\text{1})}$ and output $\mathrm{X}^{(\text{2})}$, we could get
\begin{equation}
    \begin{aligned}
        \mathrm{X}^{(\text{2})} &= (\mathrm{X}^{(\text{1})} \ast \mathrm{W}^\prime_{(b)} + \mathrm{B}^\prime_{(b)}) + (\mathrm{X}^{(\text{1})} \ast \mathrm{W}^\prime_{(n)} + \mathrm{B}^\prime_{(n)}) \\
        &= \mathrm{X}^{(\text{1})} \ast (\mathrm{W}^\prime_{(b)} + \mathrm{W}^\prime_{(n)}) + (\mathrm{B}^\prime_{(b)}) + \mathrm{B}^\prime_{(n)}) \\
        &= \mathrm{X}^{(\text{1})} \ast \mathrm{W}^\prime + \mathrm{B}^\prime. 
    \end{aligned}
   \label{Eqn:convfusion}
\end{equation} 
In the novel categories inference stage, the base and novel branches could be merged based on the additivity of convolution without any output losses.
For the next incremental learning stage, a new novel branch could be extended again and continuously merge.

\subsubsection{Adaptive Feature Fusion}
IMM could learn novel branches based on the base branch, but could not ensure the discrimination of final features for old tasks is not affected.
Here, we argue that the old categories could be seen as the unknown for the novel branch, 
thus the novel branch would output relatively lower magnitude features due to the absence of old training data~\cite{dhamija2018reducing,chen2020learning,chen2021adversarial}.
To further reduce the magnitude of the novel branch to the old classes, we propose an Adaptive Feature Fusion (AFF) mechanism to guide the learning process of the novel branches to focus on the unimportant parts of the base.
As shown in Fig.\ref{fig:structure}(b), we use the base outputs to control the propagated information through the novel branch.
The larger the output value of the base branch is, the more important its corresponding neurons are.
So we control the update of the novel branch by using negative gating functions of base outputs.
With the gated function $\mathcal{G}(.)$, the AFF with base branch $f_b(.)$ and novel branch $f_n(.)$ could be formalized as:
\begin{equation}
    \begin{aligned}
        \mathrm{X}^{(\text{2})} &= f_b(\mathrm{X}^{(\text{1})}) + g(-f_b(\mathrm{X}^{(\text{1})})) \otimes f_n(\mathrm{X}^{(\text{1})}) \\
        &= (\mathrm{X}^{(\text{1})} \ast \mathrm{W}^\prime_{(b)} + \mathrm{B}^\prime_{(b)}) \\
        &- \mathcal{G}(\mathrm{X}^{(\text{1})} \ast \mathrm{W}^\prime_{(b)} + \mathrm{B}^\prime_{(b)}) \otimes (\mathrm{X}^{(\text{1})} \ast \mathrm{W}^\prime_{(n)} + \mathrm{B}^\prime_{(n)}), \\
    \end{aligned}
    \label{Eqn:GDF}
\end{equation}
where $\mathcal{G}(.)$ is a \textit{sigmoid} function. 
Moreover, the gradient of AFF for the novel branch is $\frac{\partial \mathrm{X}^{(\text{2})}}{\partial f_n(\mathrm{X}^{(\text{1})})} = -\mathcal{G}(f_b(\mathrm{X}^{(\text{1})}))$.
Hence, the update of the novel branch is influenced by the outputs of the base branch, and the magnitude for old categories is further suppressed as shown in Fig.~\ref{fig:magnitude}.

\subsubsection{Adaptive Model Merging}
The gating process of AFF is one that varies dynamically according to the sample, which could not implement the merging operation according to the additivity of convolution, as shown in Eq.~\eqref{Eqn:GDF}.
To solve this problem, we rethink the importance of BN.
For a certain $\mathbf{BN}$ layer, $\vect{\mu}$ and $\vect{\sigma}$ compute the mean and the standard deviation, respectively, of all activations overall pixel locations for the current mini-batch data; $\vect{\gamma}$ and $\vect{\beta}$ are the trainable scaling factor and offset, respectively. 
The factor $\vect{\gamma}$ evaluates the correlation between the input and output during training. 
The gradient of the loss will approach 0 if $\vect{\gamma} \rightarrow 0$ at one training step, implying that the input will almost lose its influence on the final prediction and become redundant thereby at this training step. 
That means that if the scaling factor of one channel (with sparsity constraints) is lower than the small threshold at one training step, this channel will hardly recover and almost become redundant during the later training process.
Therefore, it motivates us to replace the input of the gated function with $\vect{\gamma}_{(b)}$ of the base $\mathbf{BN}$.
As shown in Fig.\ref{fig:structure}(c), we adopt $\vect{\gamma}_{(b)}$ of the base $\mathbf{BN}$ to control the propagated information through the novel branch, which could be formalized as:
\begin{equation}
    \begin{aligned}
        \mathrm{X}^{(\text{2})} &= f_b(\mathrm{X}^{(\text{1})}) + \mathcal{G}(-\vect{\gamma}_{(b)}) \otimes f_n(\mathrm{X}^{(\text{1})}).
    \end{aligned}
    \label{Eqn:AMM}
\end{equation}
Then, the base and novel branches could be merged based on linear additivity:
\begin{equation}
    \begin{aligned}
        \mathrm{X}^{(\text{2})} &= \mathrm{X}^{(\text{1})} \ast(\mathrm{W}^\prime_{(b)} + g(-\vect{\gamma}_{(b)}) \otimes \mathrm{W}^\prime_{(n)}) \\
        & + (\mathrm{B}^\prime_{(b)} + \mathcal{G}(-\vect{\gamma}_{(b)}) \otimes \mathrm{B}^\prime_{(n)}).   \\
    \end{aligned}
    \label{Eqn:AMMfusion}
\end{equation}
Since the base parameter is invariant during the incremental learning phase, $\vect{\gamma}_{(b)}$ corresponds to a constant, allowing merging to be implemented.
Moreover, the merging process of $\vect{\gamma}$ during base-novel merging could also be added linearly as:
\begin{equation}
    \begin{aligned}
    \vect{\gamma}^\prime = \vect{\gamma}^\prime_{(b)} + \mathcal{G}(-\vect{\gamma}_{(b)}) \otimes \vect{\gamma}^\prime_{(n)}.
    \end{aligned}
    \label{Eqn:weight}
\end{equation}
Overall, Algorithm~\ref{alg:ADM} summarizes the details of incremental category discovery with gated linear merging.

\begin{algorithm}[tbp]
  \caption{Adaptive Discovering and Merging}
  \label{alg:ADM}
  \begin{algorithmic}[1]
  \State{\bf Input:} current base model $M_b^{t}(\cdot)$, current task dataset $\mathcal{D}^u_t$, and current base weight $\vect{\gamma}_{(b)}^t$ 
        \State {\bf Novel Branch Expansion:} $M^t = \{M_b^{t}(\cdot), M_n^{t}(\cdot)\}$
         \For{all $(x_1,x_2,\dots,x_n) \in \mathcal{D}^u_t$}
            \State Compute output of $M^t$ based on Eq.\eqref{Eqn:AMM}.
            \State Compute adaptive novel class discovery loss $\mathcal{L} = \mathcal{L}^r + \mathcal{L}^c$ for mini-batch.
            \State Perform backward through $M_n^t(\cdot)$.
        \EndFor 	
        \State $\mathrm{CONV}+\mathrm{BN}$ fusion of $M_n^t(\cdot)$.
        \State {\bf Adaptive Model Merging} of base-novel parameters $\{M_b^{t}(\cdot), M_n^{t}(\cdot)\}$ based on Eq.\eqref{Eqn:AMMfusion}.
        \State {\bf Weight update} of $\vect{\gamma}_{(b)}^{t+1}$ based on Eq.\eqref{Eqn:weight}.
         \State{\bf Output:} The base model $M_b^{t+1}(\cdot)$ and weight $\vect{\gamma}_{(b)}^{t+1}$ for task $t+1$.
  \end{algorithmic}
\end{algorithm}

%% file: Main/Tex/04_experiments.tex
\begin{table*}[t]
\centering
\small
\begin{tabular}{lcccccccccccccccc}
    \toprule
     \multirow{2}{*}{\textbf{Methods}} & 
     & \multicolumn{3}{c}{\textbf{CIFAR-10}} &
     & \multicolumn{3}{c}{\textbf{CIFAR-100}} &
     & \multicolumn{3}{c}{\textbf{Tiny-ImageNet}} &
     & \multicolumn{3}{c}{\textbf{Average}} \\
     \cmidrule{3-5}\cmidrule{7-9}\cmidrule{11-13}\cmidrule{15-17}
     & 
     & \textbf{Old} & \textbf{New} & \textbf{All} &
     & \textbf{Old} & \textbf{New} & \textbf{All} &
     & \textbf{Old} & \textbf{New} & \textbf{All} &
     & \textbf{Old} & \textbf{New} & \textbf{All}  \\ 
     \midrule
     AutoNovel && 27.5 & 3.5 & 15.5 && 2.6 & 15.2 & 5.1 && 2.0 & 26.4 & 4.5 && 10.7 & 15.0 & 8.4 \\
     ResTune && 91.7 & 0.0 & 45.9 && \textbf{73.8} & 0.0 & 59.0 && 44.3 & 0.0 & 39.9 && \textbf{69.9} & 0.0 & 48.3 \\
     NCL && \textbf{92.0} & 1.1 & 46.5 && 73.6 & 10.1 & 60.9 && 0.8 & 6.5 & 1.4 && 55.5 & 5.9 & 36.3 \\
     DTC && 64.0 & 0.0 & 32.0 && 55.9 & 0.0 & 44.7 && 35.5 & 0.0 & 32.0 && 51.8 & 0.0 & 36.2 \\
     FRoST && 77.5 & 49.5 & 63.4 && 64.6 & \textbf{45.8} & 59.2 && 54.5 & 33.7 & 52.3 && 65.5 & 43.1 & 58.3 \\
     \midrule
     FRoST+\textbf{AFF} && 77.3 & 51.5 & 64.1 && 65.5 & \textbf{45.8} & 61.0 && 56.4 & 35.0 & 54.3 && 66.4 & 44.1 & 59.9 \\
     FRoST+\textbf{AMM} && 76.8 & \textbf{52.6} & 64.2 && 65.6 & 43.7 & 61.1 && 57.0 & 35.0 & 54.7 && 66.5 & 43.8 & 60.2 \\
     \textbf{ADM} && 78.4 & 51.9 & \textbf{65.2} && 67.9 & 45.7 & \black{63.2} && \textbf{57.8} & \textbf{36.0} & \textbf{55.6} && 68.0 & \textbf{44.5} & \black{61.3} \\
     \bottomrule
\end{tabular}
\caption{Comparison with state-of-the-art methods in the one-step setting of class-iNCD. All values are percentages and the best results are indicated in bold.}
\label{tab:class_incd}
\end{table*}

\section{Experiments}
\label{sec:exps}

\subsection{Experimental Setup}

\noindent \textbf{Datasets}. 
We employ three datasets to conduct experiments for class-iNCD: CIFAR-10~\cite{krizhevsky2009learning}, CIFAR-100~\cite{krizhevsky2009learning} and Tiny-ImageNet~\cite{le2015tiny}. 
The CIFAR10 and CIFAR100 datasets contain 50,000 and 10,000 32 $\times$ 32 color images for training and testing.
The Tiny-ImageNet contains 100,000 images of 200 classes downsized to 64×64 colored images.
Each dataset is split into the old and new classes following \cite{han2020automatically,zhong2021neighborhood,liu2022residual,roy2022class}. 
For CIFAR10, 5 base classes and 5 novel classes are randomly sampled. 20 novel classes are sampled for CIFAR100 and TinyImageNet and the rest are base classes.

\noindent \textbf{Evaluation Metrics}. 
We used the evaluation protocol in \cite{roy2022class} to evaluate the performance of the test data for all the classes. Three classification accuracies are reported, denoted as \textbf{Old}, \textbf{New}, and \textbf{All}. They represent the accuracy obtained from the joint classifier head on the samples of the old, new, and old+new classes, respectively. 
We used ResNet-18~\cite{he2016deep} as the backbone in all the experiments. We have adopted most of the implementation following \cite{roy2022class}.

\subsection{Results}
\label{sec:classincd}

%
%
%
%

\subsubsection{One-Step Class-iNCD}
As shown in Table~\ref{tab:class_incd}, all the NCD methods (AutoNovel~\cite{han2020automatically}, ResTune~\cite{liu2022residual}, NCL~\cite{zhong2021neighborhood}, DTC~\cite{han2019learning}) fail to achieve a good balance between the old and new classes, implying that the baseline methods can maintain the performance in old classes. 
FRoST~\cite{roy2022class} can balance the old and new categories after learning the new category without severe catastrophic forgetting. 
We first apply AFF and AMM to the SOTA class-iNCD method FRoST~\cite{roy2022class}. 
AFF could be applicable to FRoST because it only has one incremental novel discovery task to learn. 
The combination of AFF and FRoST not only further enhances the performance of the old categories, but also boosts the performance of the new categories. 
The improvement of the old categories is mainly attributed to the gating mechanism that constrains the learning parameters of the novel branch. 
Meanwhile, the gated dynamic units enable the novel branch to exploit the feature diversity of the base branch. 
Moreover, AMM replaces the dynamic gating mechanism with the $\vect{\gamma}$ weight of BN, which achieves comparable performance to AFF and even outperforms AFF in overall performance. 
This also demonstrates the effectiveness of the $\vect{\gamma}$ weight of BN as an important assessment.
Furthermore, when combining adaptive novel class discovery with AMM, we achieve the best performance in all settings.

\subsection{Further Analysis}

\begin{table}[!htbp]
\centering
\small
\begin{tabular}{lcccc}
 \toprule
   \textbf{Methods} & \textbf{Param} & \textbf{25 stages} & \textbf{10 stages} \\
 \midrule
 \rowcolor{mygray}
 iCaRL & 469K & 50.60 & 53.78  \\ 
 \rowcolor{mygray}
 iCaRL+AANets & 530K & 56.43  & 60.26 \\ 
 \rowcolor{mygray}
 iCaRL+\textbf{AMM} & 469K & \black{60.59}  & \black{62.20}  \\ 
 PODNet  & 469K &  60.72 & 63.19 \\
 PODNet+AANet & 530K & 62.31 & 64.31 \\
 PODNet+\textbf{AMM} & 469K & \black{63.45} & \black{64.32} \\
 \rowcolor{mygray}
 NECIL$\dagger$ & 11.22M & 61.01 & 61.07 \\
 \rowcolor{mygray}
 NECIL + \textbf{AMM} & 11.22M & \black{61.49} & \black{61.83}  \\
 AFC  & 497K & 63.89 & 64.98 \\
 AFC+\textbf{AMM} & 497K & \black{65.31} & \black{66.01} \\
 \rowcolor{mygray}
 DER & 112.27M & - & \black{72.81}  \\
 \rowcolor{mygray}
 FOSTER & 11.22M & 63.83 & 67.95  \\
 DyTox+$\dagger$ & 10.73M & 64.01 & 69.28 \\
 DyTox+ + \textbf{AMM} & 10.73M & \black{65.67} & \black{70.52} \\
 \bottomrule
\end{tabular}
\caption{Performance comparison with state-of-the-art class-IL methods on CIFAR100. Param is evaluated for the model in the novel inference time. $\dagger$ denotes the reproduction results based on the official code.}
\label{tab:ILcifar}
\end{table}

\noindent \textbf{The Plug-and-play Architectural Approach}
We evaluate the effectiveness of AMM in class-incremental learning with different numbers of steps. We apply AMM to five state-of-the-art methods: iCaRL~\cite{rebuffi2017icarl}, PODNet~\cite{douillard2020podnet}, NECIL~\cite{zhu2022self}, AFC~\cite{kang2022class} and DyTox~\cite{douillard2022dytox}. As shown in Table~\ref{tab:ILcifar}, our approach significantly improves the performance of all methods in all incremental stages. Moreover, the improvement increases as the number of stages grows, indicating that our method is more robust to challenging scenarios than the previous methods. Notably, DyTox~\cite{douillard2022dytox} is a class-IL method based on Transformer architecture, which can be combined with AMM. We add AMM to the \textit{Linear} layer in MLP and achieve better performance than DyTox alone.

\subsubsection{Comparison with Architectural Methods}
DER~\cite{yan2021dynamically} is a two-stage learning approach that uses a dynamically expandable representation, which increases the computation and number of parameters with the incremental stages. In contrast, AMM achieves model fusion without introducing additional parameters or requiring any training, thus avoiding performance loss. FOSTER~\cite{wang2022foster} also has base and novel models, but its model fusion relies on knowledge distillation, which could degrade the performance. NECIL~\cite{zhu2022self} uses Re-Parameterization (RP) as a combination of our IR and a prototype-based method. However, IR negatively affects the base branch without considering any base architectural prior. To address this problem, AMM adds a gating mechanism to IR to limit the influence of the novel branch on the base branch and preserve the lossless fusion property of RP. As shown in Table~\ref{tab:ILcifar}, AMM outperforms NECIL by replacing its RP with ours.

\begin{figure}[htbp]
\centering
\includegraphics[width=0.9\linewidth]{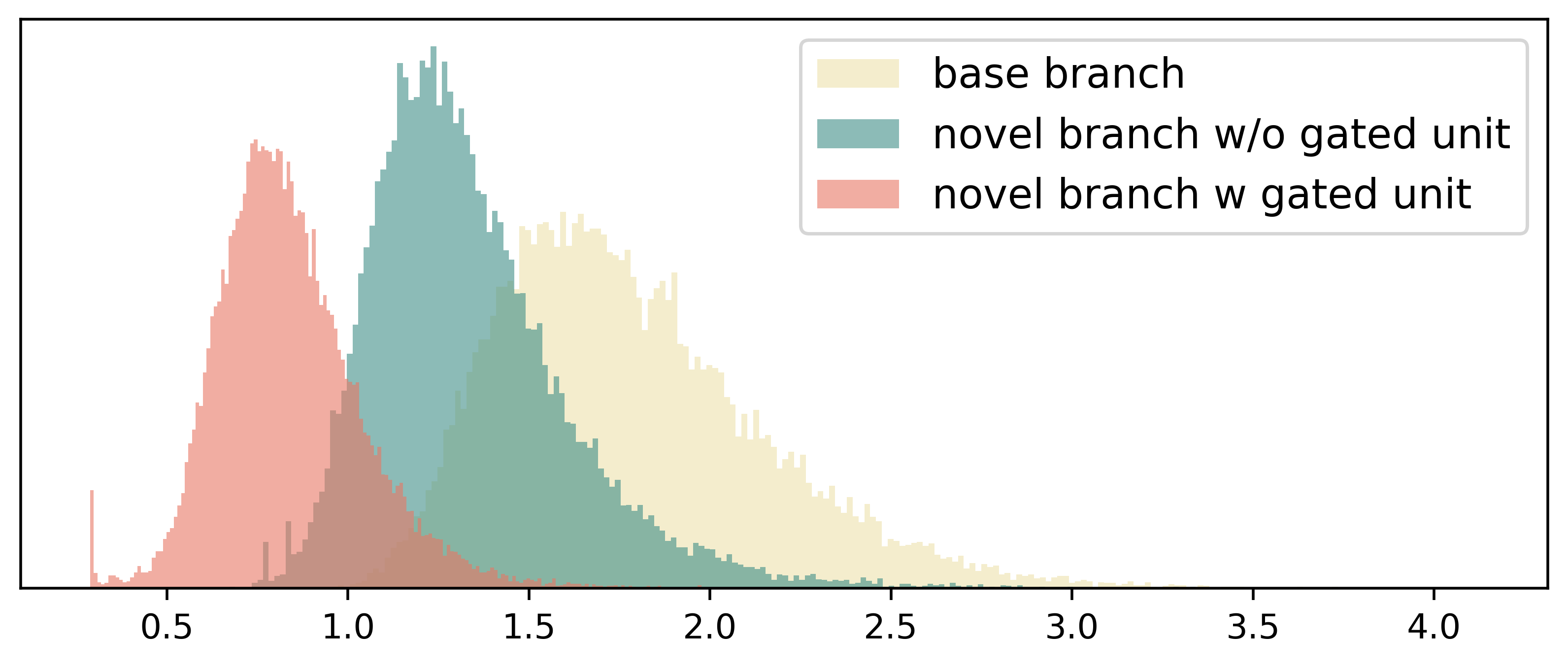}
\caption{The max magnitude distribution of old categories on the base and novel branches.}
\label{fig:magnitude}
\end{figure}

\subsubsection{The Role of the Gated Unit}
Most unknown samples have lower deep magnitude features from the neural network trained with known samples~\cite{dhamija2018reducing}, which is verified by open-set recognition~\cite{chen2020learning,chen2021adversarial}.
Therefore, the novel branch should produce low-magnitude output for old categories, to reduce the influence of old features.
As shown in Fig.~\ref{fig:magnitude}, the novel branch with the proposed gated unit achieves a lower magnitude, thus having less impact on the base branch.
Hence, the proposed AFF and AMM preserve important base features.
This also demonstrates the effectiveness of the gated unit.
The gated unit has two advantages: first, it reduces the interference of the new branch output to the important features of the base in the input phase, and leverages the discriminative power of the base features; second, it guides the novel branch to focus on the learning of unimportant parameters in the base branch, thus minimizing the influence of old categories.

%% file: Main/Tex/05_conclusion.tex
\section{Conclusion}
\label{sec:conclusion}

In this paper, we tackle two challenges in incremental novel class discovery: how to leverage novel unlabeled data sets for effective training guidance and how to prevent catastrophic forgetting of previous knowledge. We propose Triple Comparison and Probability Regularization to control the probability discrepancy and diversity of categories for adaptive category assignment. In addition, we design a hybrid structure, Adaptive Model Merging, which preserves the previous knowledge by reducing the novel branch's interference with the old classes. Extensive experiments on class-iNCD demonstrate that our method can significantly outperform the existing methods without increasing the computational cost.